\documentclass[letterpaper]{article}
\usepackage{aaai}
\usepackage{graphicx}

\usepackage{times}
\usepackage{helvet}
\usepackage{courier}

\long\def\comment#1{}
\long\def\commentout#1{}
\newcommand{\coll}[1]{\textsf{#1}}
\newcommand{\set}[1]{\textsf{#1}}

\newcommand{\eqref}[1]{Eq.~(\ref{#1})}
\newcommand{\figref}[1]{Figure~\ref{#1}}
\newcommand{\tabref}[1]{Table~\ref{#1}}
\title{Integrating Structured Metadata with Relational Affinity Propagation}
\author{Anon Plangprasopchok \and Kristina Lerman \\
USC Information Sciences Institute \\
Marina del Rey, CA\\ 
\And
Lise Getoor \\
Department of Computer Science\\
University of Maryland, College Park
}

\begin{document} 
\maketitle

\begin{abstract}
Structured and semi-structured data describing entities, taxonomies and
ontologies appears in many domains.
There is a huge interest in integrating structured information from multiple
sources; however integrating structured data to infer complex common structures is a
difficult task because the integration must aggregate similar
structures while avoiding structural inconsistencies that may appear
when the data is combined. 
In this work, we study the integration of structured social metadata: shallow personal hierarchies
specified by many individual users on the Social Web, and focus on inferring a collection of
integrated, consistent taxonomies. 
We frame this task as an optimization problem with structural constraints.
We propose a new inference algorithm, which we refer to as
\emph{Relational Affinity Propagation (RAP)} that extends affinity
propagation~\cite{apscience07} by introducing structural constraints.
We validate the approach on a real-world social media dataset,
collected from the photosharing website Flickr.  Our empirical results
show that our proposed approach is able to
construct deeper and denser structures compared to an approach using only the standard affinity
propagation algorithm.  
\end{abstract}


\commentout{
Structured data appears in many domains, including the Web, databases and bio-informatics. 
Combining structured data to learn more complex common structures is a difficult task because the integration must aggregate similar structures while avoiding structural inconsistencies that may appear once data is combined. We frame this task as \emph{an optimization with structural constraints}.

In this work, we study such optimization problem on structured social metadata: shallow personal hierarchies
specified by many individual users on the Social Web. The objective of the optimization is to combine these small
shallow hierarchies into single, or a very few, deeper and denser hierarchies that represent how a community of users organize knowledge. We propose Relational Affinity Propagation that extends Affinity Propagation by introducing novel structural constraints.

We validate the approach on the real-world dataset. The empirical results suggest that, once all inconsistencies removed after optimization, the proposed approach can combine many more personal hierarchies into deeper and denser ones,
comparing to the standard Affinity Propagation\comment{without the constraint}. \comment{Superior performance
can be achieved if combining structural constraints with the optimization process.}
}

\section{Introduction}

Structured and semi-structured data describing entities, relationships
among entities, and taxonomies and ontologies over them, appear in
many domains.  There is a great deal of interest in integrating
structured information from multiple sources.  Some of the areas that
have seen much active research include bioinformatics, aggregation
services for commercial products and services, and more traditional
enterprise database integration.  Integrating structured data to infer
complex common structures is a difficult task because the integration
must aggregate similar structures while avoiding structural
inconsistencies that may appear when the data is combined.

This problem becomes even more challenging when one is attempting to integrate
numerous, heterogeneous metadata fragments, generated by multiple users.  This
data is inherently noisy and inconsistent, and there is certainly no single, unified
structure to be found.  On the other hand, finding and extracting the best exemplar
or a set of good example structures can be highly beneficial.

In folksonomy learning~\cite{www09folksonomies}, structured metadata
in the form of hierarchies of concepts created by many users on the
Social Web is combined into a global hierarchy of concepts, that
reflects how a community organizes knowledge.  Users who create
personal hierarchies to organize their content may use idiosyncratic
categorization schemes~\cite{Golder06} and naming conventions. Simply
combining nodes with similar names is likely to lead to ill-structured
graphs containing loops and shortcuts (multiple paths from one node to
another), rather than a taxonomy.

In this paper, we present a probabilistic approach for aggregating
relational data into a desired structure.  Specifically, our task is
to integrate many shallow personal hierarchies, namely
\emph{saplings}, into a deeper, more complete taxonomy. Our learning
method, relational affinity propagation, extends affinity
propagation~\cite{apscience07} by introduces structural constraints
that encourages the integration process to combine saplings into trees
rather than an arbitrary graph containing loops and shortcuts. We show
that embedding the constraints into the hierarchy learning process
results in a more accurate merging of saplings that leads
to a more \comment{structurally} consistent tree.
We demonstrate the utility of the proposed approach on real-world data extracted from the photosharing site Flickr. Specifically, we combine shallow personal hierarchies created by Flickr users into common deeper hierarchies of concepts.

The objective of the optimization is to combine these small shallow
hierarchies into a small number of, deeper and denser
hierarchies that represent how a community of users organizes their
knowledge. This differs from classical taxonomy and ontology alignment
settings~\cite{ontomatchbook07} where there are typically just a few structures to align, and those
structures are large with rich and deep structure and semantics; here
we focus on the much messier setting, where we have many small
fragments, created by end users with a variety of purposes in mind.
In these settings, coming up with a single integrated taxonomy is infeasible, so
instead we focus on constructing a small number of useful taxonomies.

We motivate our approach with an example of learning a common taxonomy
of concepts from shallow personal hierarchies (saplings) created by
many users and illustrate some of the challenges that arise during
this task. We then briefly describe how saplings are represented
through social annotation on Flickr. We subsequently review the
standard affinity propagation algorithm and describe our relational
extension to it. Finally, we apply the method to real data sets and
show that the proposed approach is able to learn better, and more
complete trees.


\section{Motivating Example}

\begin{figure}[tb]
\begin{tabular}{cc}
\includegraphics[width=1.5in]{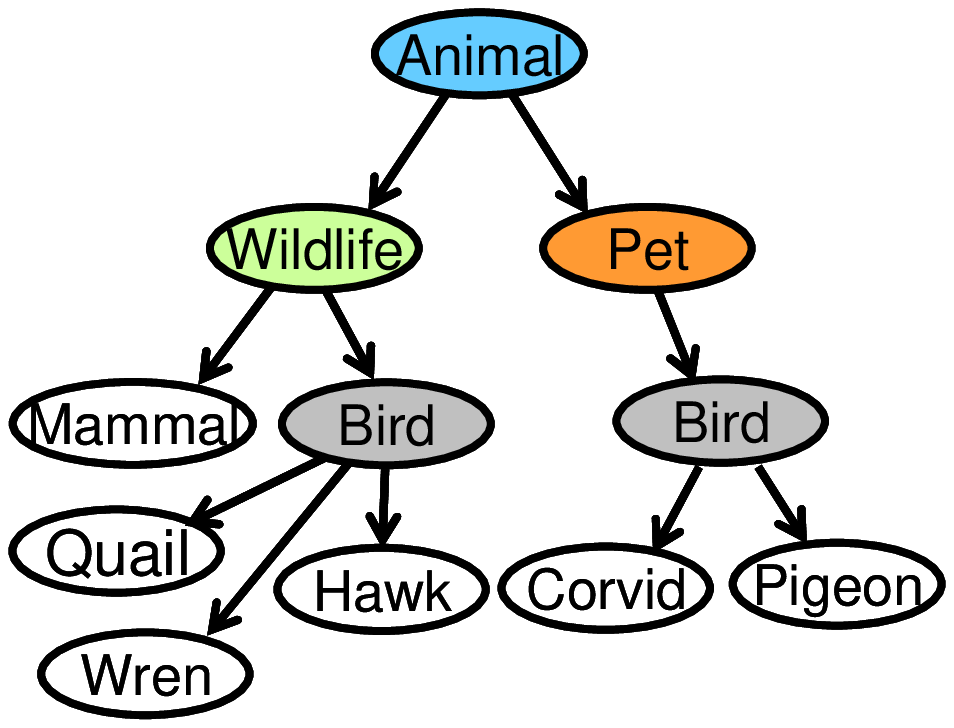}
&
\includegraphics[width=1.5in]{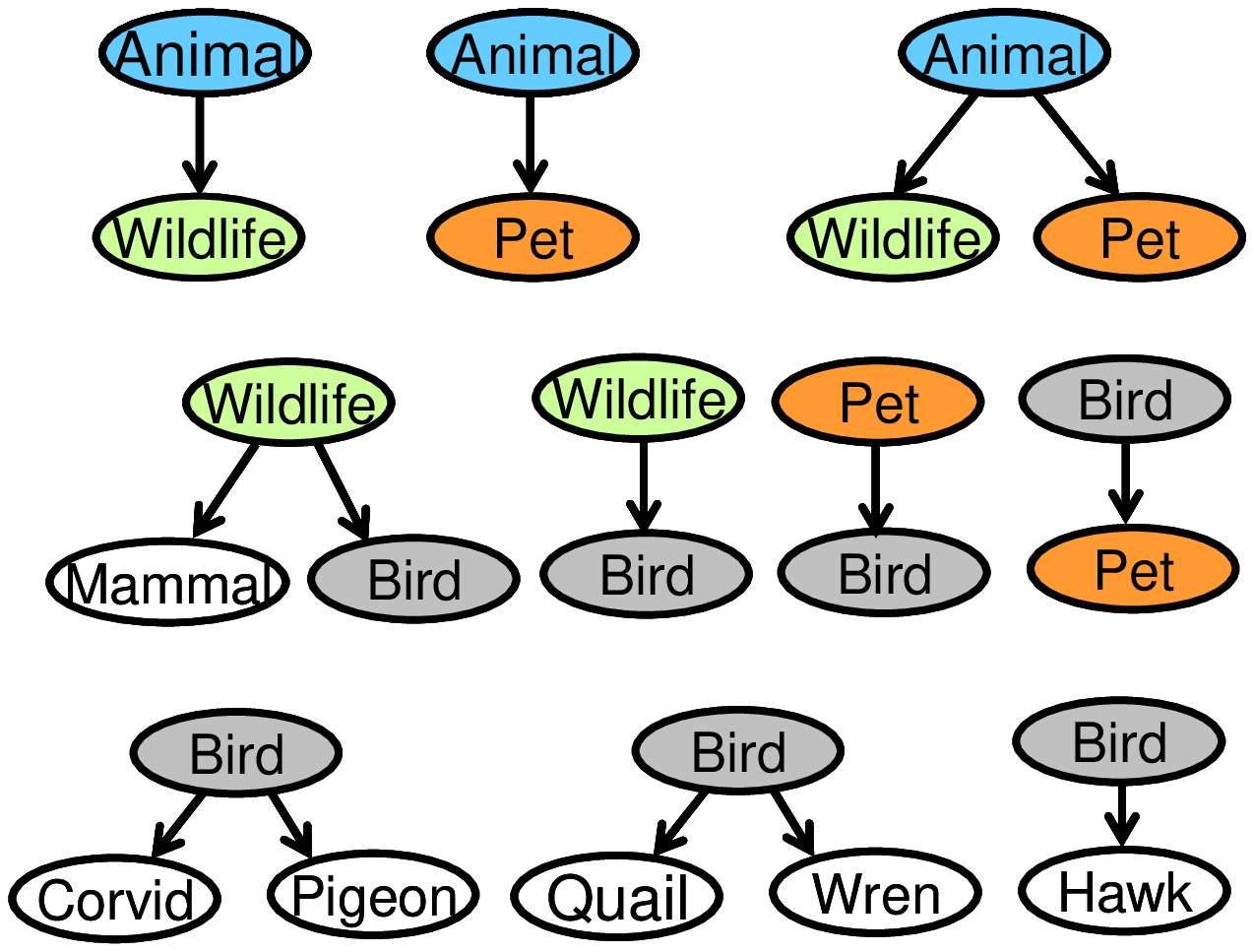} \\
(a) & (b)\\
\end{tabular}
\caption{Illustrative examples on (a) a commonly shared conceptual categorization (hierarchy) by various users;
(b) personal hierarchies expressed by the users based on the conceptual categorization as in (a). Nodes with similar name have similar color just for an illustrative purpose.}
\label{fig:motivate}
\end{figure}

We take as our motivating example user-generated annotations on the
Social Web. We assume that groups of users share common
conceptualizations of the world, which can be represented as a taxonomy
or hierarchy of concepts. \figref{fig:motivate}(a) depicts one such
common conceptualization about `animal' and its `bird' subconcepts
shared by a group of users. When users organize the content they
create, e.g., photographs on Flickr, they select some portions of the
common taxonomy for categorization. We observe these categories
through the shallow personal hierarchies Flickr users create, which 
we refer to as \emph{saplings}. \figref{fig:motivate}(b) depicts some of the
saplings specified by different users to organize their `animal' and
`bird' images. Our ultimate goal is to infer common conceptual
hierarchies from the many individual saplings. One natural solution is
to aggregate saplings shown in \figref{fig:motivate}(b) together into
a deeper and bushier tree shown in \figref{fig:motivate}(a).

To aggregate saplings, we need a 
combining strategy that measures the degree to which two sapling nodes should or should not be merged. 
Suppose that we have a very simple combining strategy that says two nodes are similar if they have similar names as in the prior work~\cite{www09folksonomies}. From \figref{fig:motivate}(b), we will end up with a graph containing one loop and two paths from `animal'
to `bird', rather than the tree shown in \figref{fig:motivate}(a). Suppose that we can also access tags with which users annotated photos within saplings, and that photos within ``domestic bird'' nodes have tags like ``pet,'' and ``farmed'' in common, and photos belonging to ``wild bird'' nodes have tags like ``wildlife'' and ``forest'' in common. A cleverer similarity function that, in addition to node names, takes tag statistics within a node into consideration, should split `bird' nodes into two different  groups: `domestic bird' and `wild bird', which are put under ``pet'' and ``wildlife'' nodes respectively.

The similarity function plays a crucial part in integrating saplings,
and a sophisticated enough similarity function that can differentiate
node senses in detail, may potentially correctly integrate the final
tree. Nevertheless, finding and tuning such function is very
difficult; moreover, the data is often inconsistent, noisy and incomplete, 
especially on the Social Web, where data is generated by many different users.

One possible way to tackle this challenge is to 
use a simple similarity function and 
incorporate constraints during the merging process. Intuitively, we would not consider merging the `bird' node under `pet' with the one under `wildlife' because it will result in multiple paths from `animal'. 
These structural constraints are used during sapling aggregation process to ensure that the learned structure is a tree. 
Specifically, the constraints prevent two nodes from being merged if (1) this will lead to links from different parent concepts or (2) this will lead to an incoming link to the root node of a tree. 
These constraints guarantee that there is, at most, a single path from one node to another.

\section{Structured Metadata in Flickr}

Structured data in the form of shallow hiearchies is ubiquitous on the
Social Web. On \emph{Flickr}, users can \emph{arbitrarily} group
related photos into \emph{sets} and then group related sets in
\emph{collections}. Some users create multi-level hierarchies
containing collections of collections, etc., but the vast majority of
users who use collections create shallow hierarchies, consisting of
collections and their constituent sets. These personal hierarchies
generally represent subclass and part-of relationships.

We formally define a \emph{sapling} as a shallow tree representing a
personal hierarchy which composed of a root node $r^{i}$ and its
children, or leaf, nodes $\langle l^{i}_{1},..l^{i}_{j} \rangle$. The
root node corresponds to a user's collection, and inherits its name,
while the leaf nodes correspond to the collection's constituent sets
and inherit their names. We assume that hierarchical relations between
a root and its children, $r^{i} \rightarrow l^{i}_{j}$, specify
broader-narrower relations.

On Flickr, users can attach tags only to photos.  A sapling's leaf node corresponds to a set of photos, and the tag statistics of the leaf are aggregated from that set's constituent photos. Tag statistics are then propagated from leaf nodes to the parent node.
\comment{In our example, \set{Plant Parasites} aggregates tag statistics from all photos in this set, and \coll{Plant Pests}, the parent of the
\set{Plant Parasites}, contains tag statistics accumulated from all photos in \set{Plant Parasites} and its siblings.}
We  define a tag statistic of node $x$ as $\tau_{x} := \{(t_{1},f_{t_{1}}), (t_{2},f_{t_{2}}), \cdots (t_{k},f_{t_{k}})\} $, where $t_{k}$ and $f_{t_{k}}$ are \emph{tag} and its frequency respectively. Hence, $\tau_{r^{i}}$ is aggregated from all $\tau_{l^{i}_{j}}$s.
These tag statistics can also be used as a feature for determining if two nodes are similar (of the same concept).


\section{Affinity Propagation} 

A key component of folksonomy learning through sapling integration is the merging similar nodes in different saplings. 
Merging similar root nodes expands the width of the learned tree, while merging the leaf of one sapling to the root of another extends the depth of the learned tree. We cast the merging process as clustering sapling nodes, and we use affinity propagation
to perform the clustering. 
Below, we briefly review the original AP and then describe our extension, which incorporates structural constraints, and refer
to the extended version as \emph{Relational Affinity Propagation (RAP)}.  

\subsection{Affinity Propagation(AP)}

Affinity Propagation~\cite{apscience07} is a clustering algorithm that
identifies a set of exemplar points that are representative of all the
points in the data set. The exemplars emerge as messages are passed
between data points, with each point assigned to an exemplar. AP attempts
to find the exemplar set which maximizes the net similarity, or the
overall sum of similarities between all exemplars and their data
points.

In this paper, we describe AP in terms of a factor graph~\cite{fg01}
on binary variables, as recently introduced by Givoni and
Frey~\cite{binAP09}. The model is comprised of a square matrix of
binary variables, along with a set of factor nodes imposed on each row
and column in the matrix. Following the notations defined in the
original paper~\cite{binAP09}, let $c_{ij}$ be a binary
variable. $c_{ij} = 1$ indicates that node $i$ belongs to node $j$
(or, $j$ is an exemplar of $i$); otherwise, $c_{ij} = 0$. Let $N$ be a
number of data points; consequently, the size of the matrix is $N
\times N$.

There are two types of constraints that enforce cluster
consistency. The first type, $I_{i}$, which is imposed on the row $i$,
indicates that a data point can belong to only one exemplar
($\sum_{j}{c_{ij}}=1$). The second type, $E_{j}$, which is imposed on
the column $j$, indicates that if a point other than $j$ chooses $j$
as its exemplar, then $j$ must be its own exemplar ($c_{jj} = 1$). AP
avoids forming exemplars and assigning cluster memberships which
violate these constraints. Particularly, if the configuration at row
$i$ violates $I$ constraint, $I_{i}$ will become $-\infty$ (and similarly for $E_{j}$).

In addition to the constraints, there is a similarity function $S(.)$, which indicates how
similar a certain node is, to its exemplar. If $c_{ij}=1$, then
$S(c_{ij})$ is a similarity between nodes $i$ and $j$; otherwise,
$S(c_{ij})=0$.  $S(c_{jj})$ evaluates ``self-similarity,'' also called
``preference'', which should be less than the maximum similarity value
in order to avoid all singleton points becoming exemplars. This is
because that configuration yields the highest net similarity.  In
general, the higher the value of the preference for a particular
point, the more likely that point will become an exemplar.  In
addition, we can set the same self-similarity value to all data
points, which indicates that all points are equally likely to be
formed as exemplars.

\begin{figure}[tb]
\begin{tabular}{cc}
\includegraphics[width=1.5in]{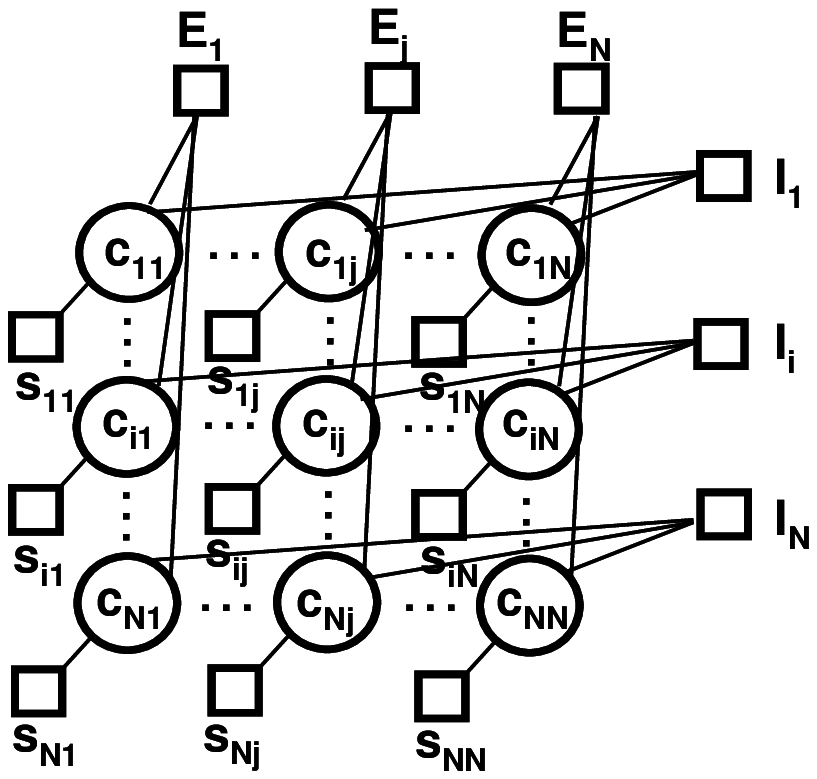}  
&
\includegraphics[width=1.5in]{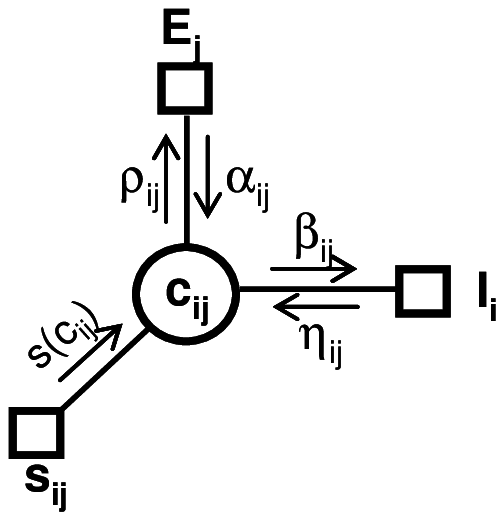} \\
(a) & (b)\\
\end{tabular}
\caption{The original binary variable model for Affinity Propagation proposed by Givoni and Frey\cite{binAP09}: (a) a matrix of binary hidden variables (circles) and their factors(boxes); (b)incoming and outgoing messages of a hidden variable node from/to its associated factor nodes.}
\label{fig:binAP}
\end{figure}

A graphical model for affinity propagation is depicted in
\figref{fig:binAP}, described in terms of a factor graph. In a
log-form, the global objective function, which measures how good the
present configuration (a set of exemplars and cluster assignments) is,
can be written as a summation of all local factors as follows:

\begin{eqnarray}
\label{eq:objBinAP} \textbf{S}(c_{11},\cdots,c_{NN}) &=&  \sum_{i,j}S_{ij}(c_{ij})+\sum_{i}I_{i}(c_{i1},\cdots,c_{iN}) \nonumber \\ 
&+&\sum_{j}E_{j}(c_{1j},\cdots,c_{1N}).
\end{eqnarray}
That is, optimizing this objective function finds the configuration that maximizes the net similarity $S$, while not violating $I$ and $E$ constraints.

The original work uses max-sum algorithm to optimize this global
objective function, and it requires updating and passing five messages
as shown in~\figref{fig:binAP}(b). Since each hidden node $c_{ij}$ is
a binary variable (two possible values), one can pass a scalar message
--- the difference between the messages when $c_{ij}=1$ and
$c_{ij}=0$, instead of carrying two messages at a time. The equations
to update these messages are described in greater detail in the
Section 2 of the original work~\cite{binAP09}.

Once the 
inference process terminates, the MAP configuration
(exemplars and their members) can be recovered as follows. First,
identify an exemplar set by considering the sum of all incoming
messages of each $c_{jj}$ (each node in the diagonal of the variable
matrix). If the sum is greater than $0$ (there is a higher probability
that node $j$ is an exemplar), $j$ is an exemplar. Once a set of
exemplars $K$ is recovered, each non-exemplar point $i$ is assigned to
the exemplar $k$ if the sum of all incoming messages of $c_{ik}$ is
the highest compared to the other exemplars.

\subsection{Relational Affinity Propagation(RAP)}

We extend the above algorithm to add in structural constraints that
will ensure that the learned folksonomy makes sense -- no loops, and,
to the extent possible, forms a taxonomy.  In fact, here, we require it
to be a tree.  Since we want the learned
folksonomy to be a tree, all nodes assigned to some exemplar must have
their incoming links from nodes in the same cluster, i.e., assigned to
the same exemplar.
To achieve this, we must enforce 
the following two constraints: (1) merging should not create incoming
links to a cluster, or concept, from more than one parent cluster
(single parent constraint); (2) merging should not create an incoming
link to the root of the induced tree (no root parent constraint). For
the second constraint, we can simply discard all sapling leaves that
are named similar to the tree root. Hence, we only need to enforce the
first constraint.  The first constraint will be violated if leaf nodes
of two saplings are merged, i.e., assigned to the same exemplar, while
the root nodes of these saplings are assigned to different exemplars.
Consequently, the leaf cluster will have multiple parents pointing to
it, which leads to an undesirable configuration.

Let $pa(.)$ be a function that returns the index of the parent node of
its argument, and $explr(.)$ be a function that return the index of
the argument's exemplar. The factor $F$, ``single parent constraint'',
checks the violation of multiple parent concepts pointing to a given
concept. The constraint is formally defined as follows:
\begin{equation}
\label{eq:fconst}
F_{j}(c_{1j},\cdots,c_{Nj}) = \left\{
\begin{array}{l l}
-\infty & \quad \mbox{$\exists i,k: c_{ij} = 1$;$c_{kj} = 1$;} \\
 & \mbox{$explr(pa(i)) \neq explr(pa(k))$,}\\
0 & \quad \mbox{otherwise.}
\end{array} \right.
\end{equation}

\begin{figure}[tb]
\begin{tabular}{cc}
\includegraphics[width=2.0in]{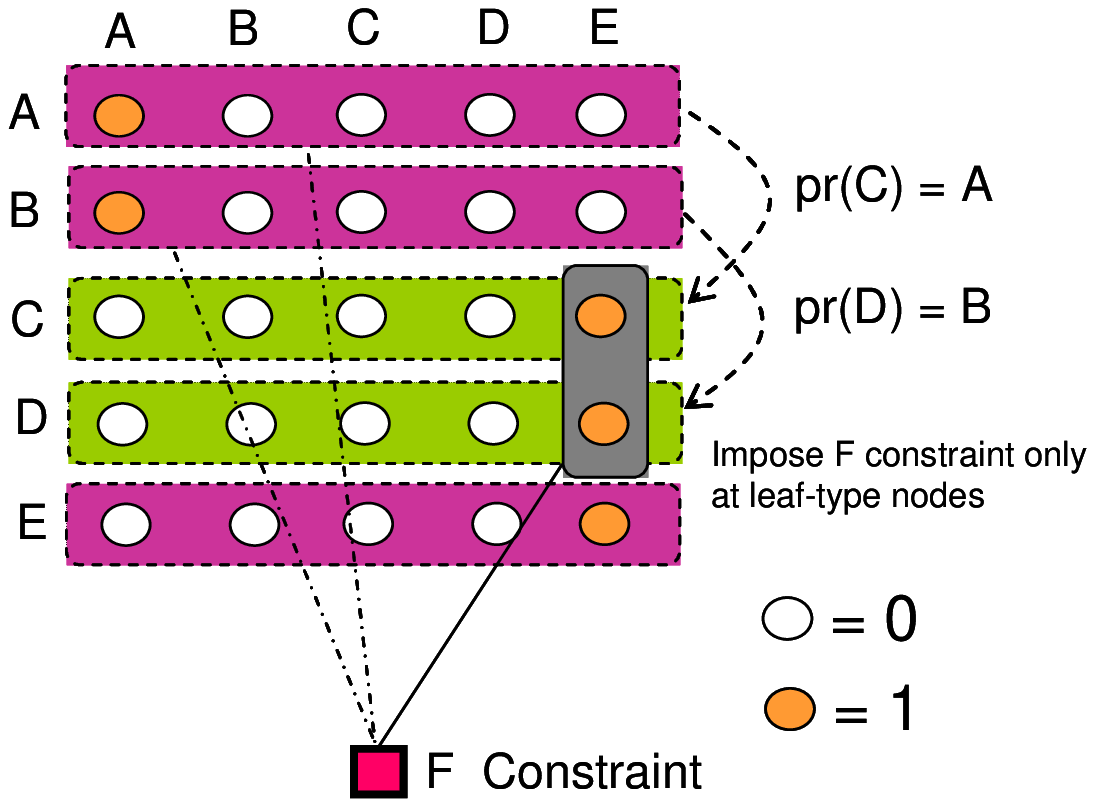}  
&
\includegraphics[width=1.25in]{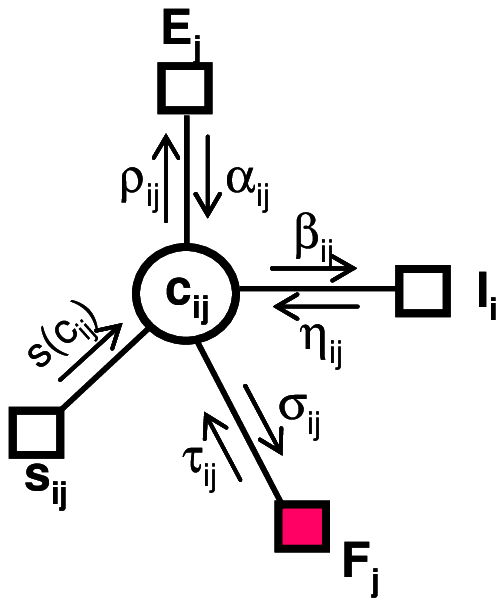} \\
(a) & (b)\\
\end{tabular}
\caption{The Relational Affinity Propagation proposed in this paper: (a) a schematic diagram of the matrix of binary hidden variables(circles), and variables within the green shade corresponding to some leaf nodes; while those within the pink shade corresponding to some root nodes of some saplings. (b) incoming and outgoing messages of a hidden variable node from/to its associated factor nodes. There are two more messages $\sigma$ and $\tau$. Note that for (a), we omit $E$, $I$ and $S$ factors simply for the sake of clarity.}
\label{fig:extAP}
\end{figure}

\figref{fig:extAP}(a) illustrates the way we impose the new constraint on the binary variable matrix. The configuration shown in the figure is valid since both $C$ and $D$ belong to the same exemplar $E$ and  their respective parents, $A$ and $B$,  belong to the same exemplar $A$. However, if $c_{BB} = 1$, then the configuration is invalid, because parents of nodes in the cluster of exemplar $E$ will belong to different exemplars.  This constraint is imposed only on leaf nodes, because merging root nodes will never lead to multiple parents. The global objective function for Relational Affinity Propagation is basically \eqref{eq:objBinAP} plus $\sum_{j}F_{j}(c_{1j},\cdots,c_{Nj})$.

We  modify the equations for updating the messages $\rho$, $\beta$ and also derive $\sigma$ and $\tau$ to take into account this additional constraint. Following the max-sum message update rule from a variable node to a factor node (cf., eq. 2.4 in Chapter 8 of~\cite{prml06}), the message update formulas for $\rho$, $\beta$ and $\sigma$ are simply:
\begin{equation}
\label{eq:extrho}
\rho_{ij} = S(i,j)+\eta_{ij}+\tau_{ij},
\end{equation}
\begin{equation}
\label{eq:extbeta}
\beta_{ij} = S(i,j)+\alpha_{ij}+\tau_{ij},
\end{equation}
\begin{equation}
\label{eq:sigma}
\sigma_{ij} = S(i,j)+\alpha_{ij}+\eta_{ij}.
\end{equation}
 
For deriving the message update equation for $\tau$, we have to consider two cases: $i = j$ and $i \neq j$, i.e., the $\tau$ message to the nodes on the diagonal and $\tau$ for the others. For simplicity, we also assume that all leaf nodes have their index numbers less than any roots. Let $L$ be a number of leaf nodes. Hence, leaf node indices run from $1$ to $L$.

For the case $i=j$ (for the diagonal nodes $c_{jj}$), we have to consider the update message for $\tau$ in two possible settings: $c_{jj}=1$ and $c_{jj}=0$ (or, they can be written as $\tau_{jj}(1)$ and $\tau_{jj}(0)$ respectively), and then find the best configuration for these settings. Following the max-sum message update rule from a factor node to a variable node (cf., eq. 2.5 in Chapter 8 of~\cite{prml06},  
when $c_{jj}=1$:
\begin{equation}
\label{eq:taujjone}
\tau_{jj}(1) = \max_{\texttt{S}^j}\{\sum_{k \in \texttt{S}^j; k \neq j}{\sigma_{kj}(1)} + \sum_{l \notin \texttt{S}^j;l \neq j}{\sigma_{lj}(0)}\}.
\end{equation}

For $c_{jj}=0$, we have 
\begin{equation}
\label{eq:taujjzero}
\tau_{jj}(0) = \sum_{k=1:L; k \neq j} \{ \sigma_{kj}(0) \},
\end{equation}
\noindent where $\texttt{S}^j \in \texttt{T}$; $\texttt{T} \supset \{1,\cdots,L\}$; $j \in \texttt{S}^j$ and all $k$ in $\texttt{S}^j$ shares the same parent exemplar. \eqref{eq:taujjone} will favor the ``valid'' configuration (the assignments of $c_{kj}$), which maximizes the summation of all incoming messages to the factor node $F_{j}$. For \eqref{eq:taujjzero}, since no other nodes can belong to $j$, the valid configuration is simply setting all $c_{kj}$ to $0$. Note that we omit $F_{j}$ from the above equations since invalid configurations are not very optimal, so that they will never be chosen. Thus, $F_{j}$ is always $0$. 

From \eqref{eq:taujjone} and \eqref{eq:taujjzero}, the scalar message $\tau_{jj}$ is simply:
\begin{equation}
\label{eq:taujj}
\tau_{jj} = \max\left\{
\begin{array}{l}
\max_{\texttt{S}^j}{\sum_{k \in \texttt{S}^j; k \neq j}{\sigma_{kj}}} \\
0
\end{array} \right.
\end{equation}

For $i \neq j$, we also have to consider two sub cases in the same way as to the previous setting,

when $c_{ij}=1$:
\begin{equation}
\label{eq:tauijone}
\tau_{ij}(1) = \max_{\texttt{S}^j}\{\sum_{k \in \texttt{S}^j; k \neq i}{\sigma_{kj}(1)} + \sum_{l \notin \texttt{S}^j;l \neq i}{\sigma_{lj}(0)}\}.
\end{equation}

For $c_{ij}=0$, we have
\begin{equation}
\label{eq:tauijzero}
\tau_{ij}(0) = \max_{\texttt{S}}\{\sum_{k \in \texttt{S}; k \neq i}{\sigma_{kj}(1)} + \sum_{l \notin \texttt{S};l \neq i}{\sigma_{lj}(0)}\},
\end{equation}

\noindent where $\texttt{S} \in \texttt{T}$; $\texttt{T} \supset \{1,\cdots,L\}$, and all $k$ in $\texttt{S}$ shares the same parent exemplar without the restriction that \texttt{S} must contain $j$. In particular, the best configuration may or may not have $j$ as the exemplar, which is different from the $c_{ij}=1$ case that requires the best configuration necessarily having $j$ as the exemplar.  

The scalar message $\tau_{ij}$, which is a difference between $\tau_{ij}(1)$ (\eqref{eq:tauijone}) and $\tau_{ij}(0)$ (\eqref{eq:tauijzero}) is as follows:
\begin{equation}
\label{eq:tauij}
\tau_{ij} = \sum_{k \in \texttt{S}^j; k \neq i}{\sigma_{kj}} - \sum_{l \notin \texttt{S};l \neq i}{\sigma_{lj}}.
\end{equation}

The inference of exemplars and cluster assignments starts by
initializing all messages to zero and keeps updating all messages for
each nodes iteratively until convergence. One possible way to determine
the convergence is to monitor the stability of the net similarity
value, $\sum_{i,j}S_{ij}(c_{ij})$, as in the original AP.

Recovering MAP exemplars and cluster assignments can be done as in the
original AP with one extra step, in order to guarantee that the final
graph is in a tree form. In particular, for a certain exemplar, we
sort its members by their message summation value in descending
order. Note that the higher the value, the more likely the node
belongs to its exemplar. The parent exemplar of a cluster of nodes is
determined as follows. If the exemplar of the cluster is a leaf node,
the parent exemplar of the cluster is the parent exemplar of the
exemplar. Otherwise, the parent exemplar of the highest-ranked leaf
node will be chosen. We then split all member nodes that have
different parent exemplars to that of the cluster. Note that a more
sophisticated approach to this task may be applied: e.g., once
split, find the next best valid exemplar to join. However, this more
complex procedure is very cumbersome -- the decision to re-join
a certain cluster may recursively result in the invalidity of 
other clusters.


\section{Validation on a Toy Example}

To evaluate the utility of RAP, we first apply it to a simplified data
set, which consists of a small fraction of the personal hierarchies
taken from the Flickr data set described
in~\cite{www09folksonomies}. These hierarchies are about `animal',
`pet', `wildlife', `bird' (wild and domestic), which are very similar
to \figref{fig:motivate}(b). The ideal integrated hierarchy is similar
to \figref{fig:motivate}(a), where `bird' concept is split into
domestic and wildlife birds under the `animal' concept. There are
total of $96$ saplings generated by different users in this data set.

We quantitatively compare the quality of the tree learned by RAP
against that learned by the standard AP algorithm. In addition, since the AP
does not have machinery for ``correcting'' the output graph into a
tree, after the final inference step, we run the same procedure that
recovers exemplars and valid cluster assignments that is used in the
final step of RAP. We used the following evaluation metrics: net
similarity and tree depth and bushiness. Intuitively, we prefer ``a
tree of exemplars,'' which clusters as many similar nodes as possible
(high net similarity), as well as a comprehensive tree (bushy and
deep).

We used a simple similarity function, $S(i,j)$, to compute the
similarity between two nodes $i$ and $j$. Let $t^{ij}$ be a number of
common tags of $i$ and $j$ nodes. If $i$ and $j$ have the same stemmed
name, $S(i,j) = min(1.0,t^{ij})$ (if they have, at least, just one tag
in common, the similarity value goes to 1); otherwise, $0$. The
damping factor is set to $0.9$, and the number of iterations is set
to $4,000$. The preference is set to $0.0001$ uniformly.

The inference converges in both approaches before $4,000$ iterations
and returns a single `animal' tree. The net similarities of the trees
(after correcting the graphs) are $83.41$ (AP) and $106.31$
(RAP). Both approaches return trees of similar depth, namely $3$. The
distribution of the number of exemplars at depths $\langle 0,1,2,3
\rangle$ for AP and RAP are $\langle 1,25,30,9 \rangle$ and $\langle
1,2,32,12 \rangle$ respectively, and the distribution of the number of
instances (data points) at depths $\langle 0,1,2,3 \rangle$ are
$\langle 35,78,46,10 \rangle$ for AP, and $\langle 35,81,51,13
\rangle$ for RAP. Although AP yields a ``bushier'' tree of exemplars,
this does not really demonstrate its superiority to RAP. In fact, at
the depth $1$, AP shatters the `pet' concept into many singleton
clusters; while RAP nicely merges them into a single cluster. The
distribution of the number of instances at different depths also
indicates that RAP can aggregate more nodes into a tree, compared to
AP.
Hence, RAP's overall quality on this small example is higher.

\begin{figure}[tb]
\begin{tabular}{c}
\includegraphics[width=2.4in]{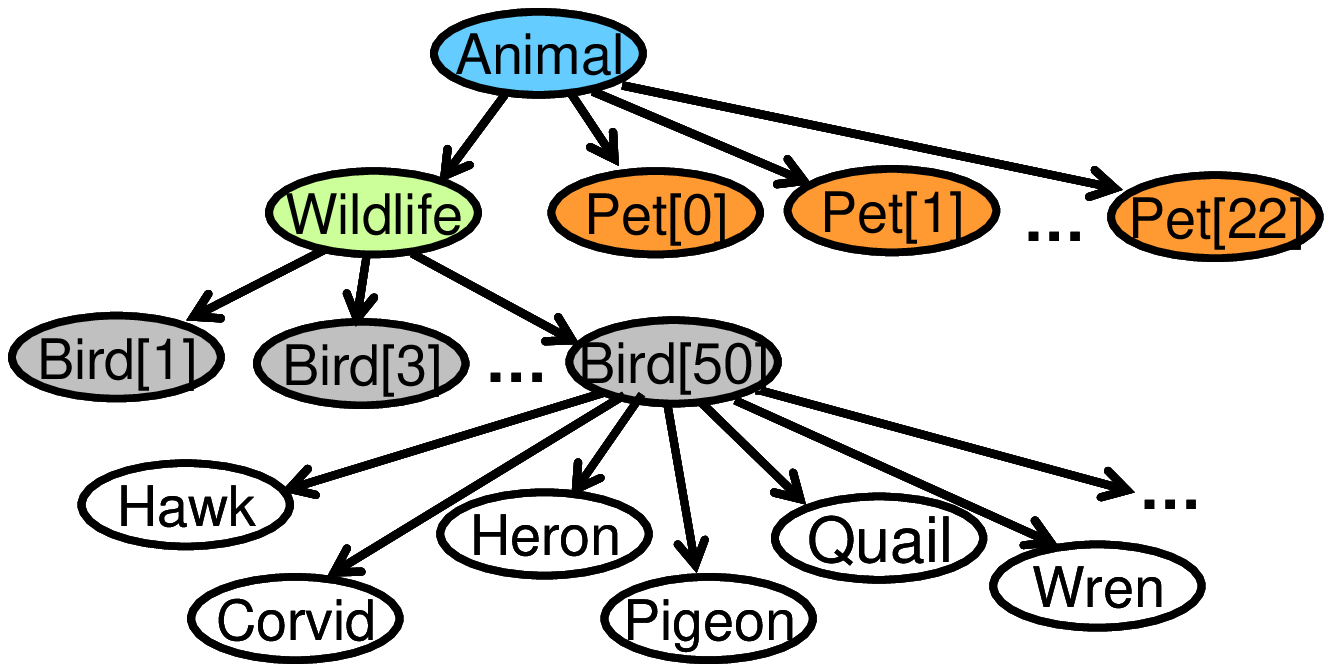}\\
(a) AP's tree \\
\includegraphics[width=2.7in]{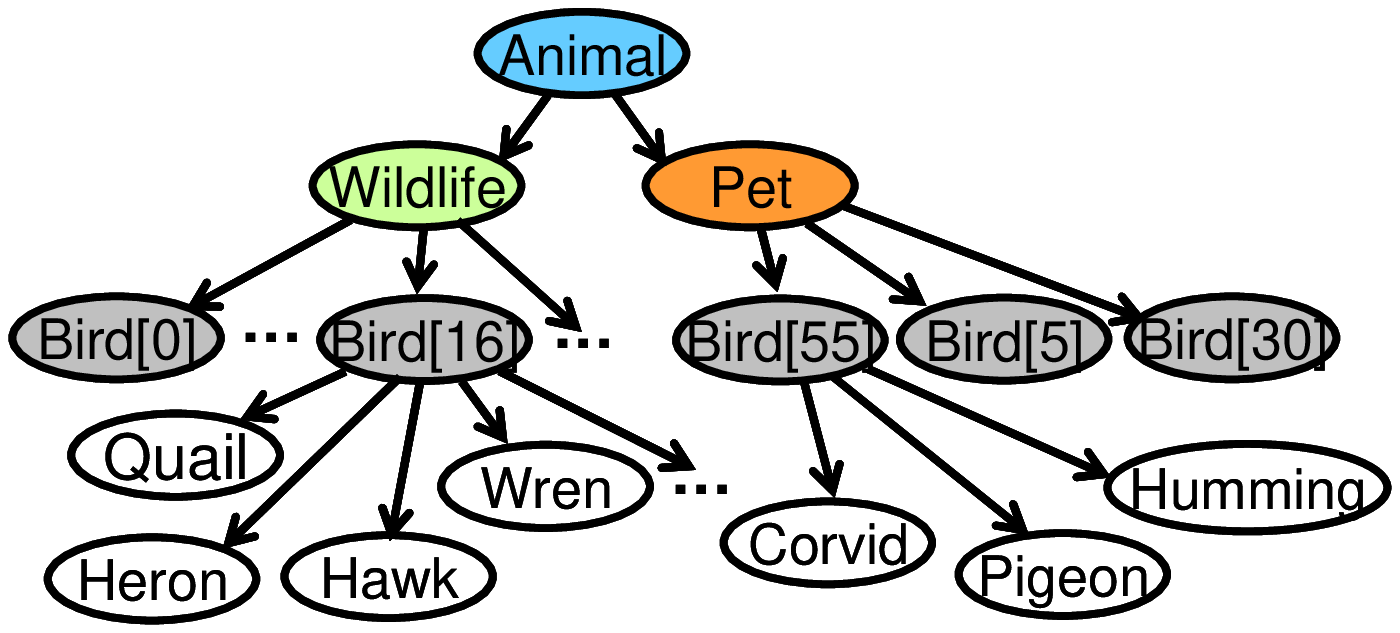} \\
(b) RAP's tree
\end{tabular}
\caption{trees induced by AP(a) and RAP(b). The numbers in brackets just indicate the index numbers of the exemplar nodes of their clusters. Non-exemplar nodes are collapsed to one of these exemplars. The nodes without number are those that cluster all nodes with the same name together.}
\label{fig:toytrees}
\end{figure}

The trees learned by both methods are shown in \figref{fig:toytrees}.
AP clusters both wild and domestic `bird' nodes together in cluster number $50$ as illustrated in \figref{fig:toytrees}(a), while, RAP separates them into two different clusters: cluster number $16$ (wild bird) and cluster number $55$ (domestic bird). By taking structural constraints into account, RAP is able to separate `bird' into different senses.
Specifically, during the inference process this constraint prevents all bird nodes from being clustered together, because this will create paths to this cluster from two different parent clusters: `wildlife' and `pet'. The only valid configuration then is to have two `bird' clusters: one under `wildlife' and one under `pet'. The inference process optimizes the tree within this valid configuration.
AP, on the other hand, does not have this machinery and, consequently, merges all `bird' nodes together without concern for where the incoming links come from. The final correction step does not optimize the tree structure, and the tree learned by AP is worse than one learned by RAP.

\section{Validation on Real-Wold Data}

We also compared RAP against AP on the data collected from
Flickr~\cite{www09folksonomies}. We manually selected $15$ seed terms,
and for each term used the following heuristic to obtain ``relevant''
saplings. First, we selected saplings whose root names were similar to
the seed term. We then used the leaf node names of these saplings to
select other saplings whose root names were similar to these names,
and so on, for the total of two iterations.  We used the settings
described above but with the number of iterations limited to $2000$.

For each seed, we ran AP and RAP on all extracted saplings; then
measured the net similarity of the induced tree. For measuring the
induced tree's structure in terms of bushiness and depth, we introduce
a simple, yet intuitive measure, namely Area Under Tree(AUT), which
takes \emph{both} tree bushiness and depth into account. To calculate
AUT for a given tree, we plot the distribution of the number
of nodes at each level and then compute the area under the
plot. Intuitively, trees that keep branching out at each level will
get a larger AUT than those that short and thin. Suppose that we have
a tree in which the number of nodes at $1^{st}$ and $2^{rd}$ level are 
$3$ and $4$, respectively. With the scale of tree depth set to $1.0$, AUT of this tree
would be $0.5\times(1+3) + 0.5\times(3+4) = 5.5$.

\begin{table}[tbh]
\centering
\setlength{\tabcolsep}{1pt}
\scriptsize{

\begin{tabular}{|lllllllll}
\hline
\multicolumn{1}{|l|}{} & \multicolumn{2}{c|}{Entire Set} & \multicolumn{6}{c|}{Induced Seed Trees} \\
\cline{2-9}
\multicolumn{1}{|l|}{} & \multicolumn{2}{c|}{Net Sim} & \multicolumn{2}{c|}{\# Trees} & \multicolumn{2}{c|}{Best Tree's Net Sim} & \multicolumn{2}{c|}{Best Tree's AUT} \\
\cline{2-9}
\multicolumn{1}{|l|}{} & \multicolumn{1}{l|}{AP} & \multicolumn{1}{l|}{RAP} & \multicolumn{1}{l|}{AP} & \multicolumn{1}{l|}{RAP} & \multicolumn{1}{l|}{AP} & \multicolumn{1}{l|}{RAP} & \multicolumn{1}{l|}{AP} & \multicolumn{1}{l|}{RAP} \\
\hline
\multicolumn{1}{|l|}{africa} & \multicolumn{1}{l|}{46.93} & \multicolumn{1}{l|}{103.92} & \multicolumn{1}{l|}{2} & \multicolumn{1}{l|}{2} & \multicolumn{1}{l|}{25.41} & \multicolumn{1}{l|}{63.31} & \multicolumn{1}{l|}{55} & \multicolumn{1}{l|}{103} \\
\hline
\multicolumn{1}{|l|}{animal} & \multicolumn{1}{l|}{3609.32} & \multicolumn{1}{l|}{3839.88} & \multicolumn{1}{l|}{4} & \multicolumn{1}{l|}{4} & \multicolumn{1}{l|}{142.48} & \multicolumn{1}{l|}{156.28} & \multicolumn{1}{l|}{606} & \multicolumn{1}{l|}{727} \\
\hline
\multicolumn{1}{|l|}{asia} & \multicolumn{1}{l|}{474.06} & \multicolumn{1}{l|}{617.15} & \multicolumn{1}{l|}{2} & \multicolumn{1}{l|}{2} & \multicolumn{1}{l|}{143.75} & \multicolumn{1}{l|}{219.86} & \multicolumn{1}{l|}{445.5} & \multicolumn{1}{l|}{523.5} \\
\hline
\multicolumn{1}{|l|}{australia} & \multicolumn{1}{l|}{167.65} & \multicolumn{1}{l|}{227.24} & \multicolumn{1}{l|}{2} & \multicolumn{1}{l|}{2} & \multicolumn{1}{l|}{72.11} & \multicolumn{1}{l|}{123.41} & \multicolumn{1}{l|}{104} & \multicolumn{1}{l|}{151} \\
\hline
\multicolumn{1}{|l|}{bird} & \multicolumn{1}{l|}{231.73} & \multicolumn{1}{l|}{365.02} & \multicolumn{1}{l|}{3} & \multicolumn{1}{l|}{3} & \multicolumn{1}{l|}{31.11} & \multicolumn{1}{l|}{33.31} & \multicolumn{1}{l|}{75} & \multicolumn{1}{l|}{71.5} \\
\hline
\multicolumn{1}{|l|}{canada} & \multicolumn{1}{l|}{278.7} & \multicolumn{1}{l|}{312.29} & \multicolumn{1}{l|}{2} & \multicolumn{1}{l|}{2} & \multicolumn{1}{l|}{46.32} & \multicolumn{1}{l|}{50.62} & \multicolumn{1}{l|}{127} & \multicolumn{1}{l|}{138} \\
\hline
\multicolumn{1}{|l|}{craft} & \multicolumn{1}{l|}{235.65} & \multicolumn{1}{l|}{285.75} & \multicolumn{1}{l|}{6} & \multicolumn{1}{l|}{6} & \multicolumn{1}{l|}{24.31} & \multicolumn{1}{l|}{24.41} & \multicolumn{1}{l|}{72.5} & \multicolumn{1}{l|}{67.5} \\
\hline
\multicolumn{1}{|l|}{fish} & \multicolumn{1}{l|}{124.83} & \multicolumn{1}{l|}{132.43} & \multicolumn{1}{l|}{1} & \multicolumn{1}{l|}{1} & \multicolumn{1}{l|}{44.61} & \multicolumn{1}{l|}{45.71} & \multicolumn{1}{l|}{69} & \multicolumn{1}{l|}{68.5} \\
\hline
\multicolumn{1}{|l|}{insect} & \multicolumn{1}{l|}{244.69} & \multicolumn{1}{l|}{257.19} & \multicolumn{1}{l|}{31} & \multicolumn{1}{l|}{31} & \multicolumn{1}{l|}{34.81} & \multicolumn{1}{l|}{37.1} & \multicolumn{1}{l|}{66.5} & \multicolumn{1}{l|}{63} \\
\hline
\multicolumn{1}{|l|}{invertebrate} & \multicolumn{1}{l|}{34.01} & \multicolumn{1}{l|}{56.2} & \multicolumn{1}{l|}{1} & \multicolumn{1}{l|}{1} & \multicolumn{1}{l|}{32.91} & \multicolumn{1}{l|}{55.11} & \multicolumn{1}{l|}{97.5} & \multicolumn{1}{l|}{99.5} \\
\hline
\multicolumn{1}{|l|}{mammal} & \multicolumn{1}{l|}{97.86} & \multicolumn{1}{l|}{117.25} & \multicolumn{1}{l|}{2} & \multicolumn{1}{l|}{2} & \multicolumn{1}{l|}{50.71} & \multicolumn{1}{l|}{36.71} & \multicolumn{1}{l|}{58.5} & \multicolumn{1}{l|}{65} \\
\hline
\multicolumn{1}{|l|}{plant} & \multicolumn{1}{l|}{565.72} & \multicolumn{1}{l|}{714.7} & \multicolumn{1}{l|}{2} & \multicolumn{1}{l|}{2} & \multicolumn{1}{l|}{12.3} & \multicolumn{1}{l|}{13.4} & \multicolumn{1}{l|}{19.5} & \multicolumn{1}{l|}{19} \\
\hline
\multicolumn{1}{|l|}{sport} & \multicolumn{1}{l|}{725.81} & \multicolumn{1}{l|}{758.2} & \multicolumn{1}{l|}{9} & \multicolumn{1}{l|}{9} & \multicolumn{1}{l|}{92.85} & \multicolumn{1}{l|}{105.74} & \multicolumn{1}{l|}{269} & \multicolumn{1}{l|}{252.5} \\
\hline
\multicolumn{1}{|l|}{uk} & \multicolumn{1}{l|}{640.48} & \multicolumn{1}{l|}{754.16} & \multicolumn{1}{l|}{1} & \multicolumn{1}{l|}{1} & \multicolumn{1}{l|}{370.84} & \multicolumn{1}{l|}{526.7} & \multicolumn{1}{l|}{633} & \multicolumn{1}{l|}{673.5} \\
\hline
\multicolumn{1}{|l|}{usa} & \multicolumn{1}{l|}{286.28} & \multicolumn{1}{l|}{390.56} & \multicolumn{1}{l|}{1} & \multicolumn{1}{l|}{1} & \multicolumn{1}{l|}{244.29} & \multicolumn{1}{l|}{341.99} & \multicolumn{1}{l|}{354.5} & \multicolumn{1}{l|}{370} \\
\hline
\end{tabular}

}
\label{tbl:results}
\caption{The table presents the performance comparisons between AP and RAP by the net similarity for the entire data and the best induced tree (with highest net similarity) on $15$ different seed sets. The number of induced trees and Area Under Tree (AUT) are also reported.}
\end{table}

Both quantitative and manual inspections confirm the advantages of RAP
over AP. As shown in~\tabref{tbl:results}, RAP yields better net
similarity in all cases. Although both approaches return the same
number of trees,
RAP appears to better cluster  similar nodes in all but one case, namely `mammal'.
In terms of AUT, though many trees are of similar quality, in cases where significant differences exist, they are in RAP's favor.
Manual inspection reveals that AP tends to ``shatter'' trees into isolated singletons rather than merge similar nodes together, as RAP does.


\section{Related Work}
\commentout{
There are many work on learning structured from data such as learning the structure of probabilistic networks~\cite{learningbayesnet95}; perhaps, the closest ones to ours are, e.g., learning systems of concepts~\cite{IRM06}, and learning hierarchical topics from words in documents~\cite{hLDA03}\comment{, in which their common objective is to learn (hierarchical) relations between concepts}. Nevertheless, our work is fundamentally different from them in that, we ``align'' or integrate many small shallow hierarchies, whose hierarchical relations between concepts are \emph{explicitly} specified by users. Consequently, our work attempts to find the best ``alignment'', which maximizes the similarity between similar concepts, as having no structural inconsistencies.
}
Affinity propagation has been applied to many clustering problems,
e.g. segmentation in computer vision~\cite{apvision09}.  It
provides a natural way to incorporate constraints while simultaneously
improving the net similarity of the cluster assignments, which is not
trivial to handle in standard clustering techniques. In addition, no
strong assumption is required on the threshold, which determines
whether clusters should be merged or not. Moreover, the cluster
assignments can be changed during the inference process as suggested
by the emergence of exemplars.
Nevertheless, to our knowledge, there is no extension of AP algorithm
to learn tree structures from many sparse and shallow trees as
presented in this work.

There are many other SRL approaches that are applicable as well.  For
example, Markov Logic Networks (MLN)~\cite{mln06}, a generic framework
for solving probabilistic inference problems, may also be applied to
folksonomy learning, by translating similarity function as well as
constraints into predicates. Since our similarity function is
continuous, hybrid MLN (HMLN)~\cite{hmln08} would be required.  The AP
framework has advantages due to its simplicity; however we plan to
investigate more comparative work with existing SRL approaches,
especially as we explore more complex similarity functions.


\commentout{
Affinity Propagation has been applied to many clustering problems, e.g. segmentation in computer visions~\cite{apvision09}, because  it provides a natural way to incorporate constraints while simultaneously improving the net similarity of the cluster assignments, which is not trivial to handle in standard clustering techniques. In addition, no strong assumption is required on the threshold, which determines whether clusters should be merged or not. Moreover, the cluster assignments can be changed during the inference process as suggested by the emergence of exemplars, comparing to many ``incremental'' clustering (e.g.,~\cite{EntityResoluteGetoor07}), at which the previous clustering decision cannot be changed. Nevertheless, to our knowledge, there is no extension of AP algorithm to learn tree structures from many sparse and shallow trees as presented in this work.
}

\section{Discussion and Conclusion}


In this paper, we introduce relational affinity propagation (RAP), an
extension to affinity propagation to learn structures from data by
incorporating structural constraints. RAP optimizes the net similarity
and it uses the structural constraint to find good solutions within a
space of ``valid'' solutions. Thus, the final net similarity of the
tree learned by RAP is better than AP. Our validations on toy and
real-world data support this claim.  For the future work, we would
like to apply more sophisticated similarity function, which utilizes
class labels as in a collective relational clustering
approach
in order to improve the
quality of the learned structures. In addition, the structural constraints
can be modified to guide RAP to induce other classes of graphs, e.g.,
a DAG. We would also like to extend RAP to apply on other structure
learning problems.

\commentout{
In this paper, we presented relational Affinity Propagation (RAP), a novel extension to Affinity Propagation, to learn structures from data by incorporating a structural constraint. RAP optimizes the net similarity and it uses the global constraint to find good solutions within a space of ``valid'' solutions. Thus, the final net similarity of the tree learned by RAP is better than AP. Our validations on toy and real-world data supports this claim.

For the future work, we would like to apply more sophisticated similarity function, which utilizes class labels as in collective relational clustering approach~\cite{EntityResoluteGetoor07}, in order to improve the quality of the learned structures. In addition, the global constraint can be modified to guide RAP to induce other classes of graphs, e.g., DAG. We would also like to extend RAP to apply on other structure learning problems, appearing in many domains.
}

\vspace{4mm}
\noindent\textbf{Acknowledgments}: this material is based upon work supported by the National Science Foundation under Grant No.IIS-0812677.

\begin{scriptsize}

\end{scriptsize}

\begin{thebibliography}{}

\bibitem[\protect\citeauthoryear{Bishop}{2006}]{prml06}
Bishop, C.~M.
\newblock 2006.
\newblock {\em Pattern Recognition and Machine Learning}.
\newblock Springer-Verlag.

\bibitem[\protect\citeauthoryear{Euzenat and Shvaiko}{2007}]{ontomatchbook07}
Euzenat, J., and Shvaiko, P.
\newblock 2007.
\newblock {\em Ontology Matching}.
\newblock Springer-Verlag.

\bibitem[\protect\citeauthoryear{Frey and Dueck}{2007}]{apscience07}
Frey, B.~J., and Dueck, D.
\newblock 2007.
\newblock Clustering by passing messages between data points.
\newblock {\em Science} 312:972–--976.

\bibitem[\protect\citeauthoryear{Givoni and Frey}{2009}]{binAP09}
Givoni, I.~E., and Frey, B.~J.
\newblock 2009.
\newblock A binary variable model for affinity propagation.
\newblock {\em Neural Comput} 21(6):1589--1600.

\bibitem[\protect\citeauthoryear{Golder and Huberman}{2006}]{Golder06}
Golder, S.~A., and Huberman, B.~A.
\newblock 2006.
\newblock Usage patterns of collaborative tagging systems.
\newblock {\em J. Inf. Sci.} 32(2):198--208.

\bibitem[\protect\citeauthoryear{Kschischang, Frey, and Loeliger}{2001}]{fg01}
Kschischang, F.; Frey, B.~J.; and Loeliger, H.-A.
\newblock 2001.
\newblock Factor graphs and the sum-product algorithm.
\newblock {\em IEEE Transactions on Information Theory} 47:498--519.

\bibitem[\protect\citeauthoryear{Lazic \bgroup et al.\egroup
  }{2009}]{apvision09}
Lazic, N.; Givoni, I.; Frey, B.; and Aarabi, P.
\newblock 2009.
\newblock Floss: Facility location for subspace segmentation.
\newblock In {\em Proceedings of the International Conference on Computer
  Vision}.

\bibitem[\protect\citeauthoryear{Plangprasopchok and
  Lerman}{2009}]{www09folksonomies}
Plangprasopchok, A., and Lerman, K.
\newblock 2009.
\newblock Constructing folksonomies from user-specified relations on flickr.
\newblock In {\em Proceedings of the World Wide Web conference}.

\bibitem[\protect\citeauthoryear{Richardson and Domingos}{2006}]{mln06}
Richardson, M., and Domingos, P.
\newblock 2006.
\newblock Markov logic networks.
\newblock {\em Mach. Learn.} 62:107--136.

\bibitem[\protect\citeauthoryear{Wang and Domingos}{2008}]{hmln08}
Wang, J., and Domingos, P.
\newblock 2008.
\newblock Hybrid markov logic networks.
\newblock In {\em Proceedings of Association for the Advancement of Artificial
  Intelligence}.

\end{thebibliography}
\end{document}